\documentclass[conference]{IEEEtran}
\IEEEoverridecommandlockouts
\usepackage{placeins}
\usepackage{cite}
\usepackage{amsmath,amssymb,amsfonts}
\usepackage{algorithmic}
\usepackage{graphicx}
\usepackage{textcomp}
\usepackage[table]{xcolor}
\usepackage{longtable}
\usepackage{booktabs}
\usepackage{multirow}
\usepackage{array}
\usepackage{caption}
\usepackage{etoolbox}
\AfterEndEnvironment{equation}{\noindent}

\usepackage{titlesec}

\titlespacing*{\section}
{0pt}
{4pt plus 1pt minus 1pt}
{3pt plus 1pt minus 1pt}

\titlespacing*{\subsection}
{0pt}
{3pt plus 1pt minus 1pt}
{2pt plus 1pt minus 1pt}

\def\BibTeX{{\rm B\kern-.05em{\sc i\kern-.025em b}\kern-.08em
    T\kern-.1667em\lower.7ex\hbox{E}\kern-.125emX}}
\begin{document}
\newcommand{\red}[1]{\textcolor{red}{#1}}
%\title{Enhancing Day-ahead Load Forecasting using Deep Residual Learning and Multi-Modal Cross-Attention\\
\title{IMPACT: Inverter-Informed Multimodal Learning for Photovoltaic Generation Forecasting\\
\thanks{This work was authored in part by the National Laboratory (NLR) of the Rockies for the U.S. Department of Energy (DOE), operated under Contract No. DE-AC36-08GO28308. Funding for the NLR authors provided by the U.S. Department of Energy Office of Critical Minerals and Energy Innovation. The views expressed in the article do not necessarily represent the views of the DOE or the U.S. Government. The U.S. Government retains and the publisher, by accepting the article for publication, acknowledges that the U.S. Government retains a nonexclusive, paid-up, irrevocable, worldwide license to publish or reproduce the published form of this work, or allow others to do so, for U.S. Government purposes.
}
%\thanks{The views expressed in the article do not necessarily represent the views of the DOE or the U.S. Government. The U.S. Government retains and the publisher, by accepting the article for publication, acknowledges that the U.S. Government retains a nonexclusive, paid-up, irrevocable, worldwide license to publish or reproduce the published form of this work, or allow others to do so, for U.S. Government purposes.}
%\thanks{Disclaimer: The views expressed in this paper are solely those of the authors and do not necessarily represent those of MISO}
}

\author{\IEEEauthorblockN{Anushka Katiyar, Cong Feng, Jing Wang}
\textit{National Laboratory of the Rockies}\\
Golden, USA \\
\{anushka.katiyar, cong.feng, jing.wang\}@nlr.gov
\and
\IEEEauthorblockN{Tristan Liu, Simon Julien, Zachary Jacobs}
\textit{Latimer Controls}\\
Denver, USA \\
\{tristan, simon, zachary\}@latimercontrols.com
}

\maketitle

\begin{abstract}
This work develops a physics-informed multimodal deep learning framework for 10- to 110-minute-ahead probabilistic photovoltaic power forecasting. The forecasting model is built on a four-branch architecture that separately encodes plant-level temporal history and inverter-level spatial cloud dynamics through a graph attention encoder with a learned adjacency, known-future solar geometry, and the current meteorological state, fused through gated cross-attention. The geometry branch adds two forward-looking physical signals, spatial cloud heterogeneity and inverter ramp-down activity, each empirically weighted toward the horizon where it is most predictive. The decoder anchors its prediction to a smart persistence baseline, learning a horizon-conditioned correction, and the resulting quantile forecasts are calibrated online through a kernel-weighted adaptive conformal calibration step. The proposed model outperforms all 14 baselines, achieving a forecast skill score (FSS) of 0.2950, 0.099 above the next-best model. A branch-composition ablation shows that every branch contributes positively and the model retains substantial skill in challenging overcast and variable sky conditions (FSS 0.198 and 0.255).
\end{abstract}


\section{Introduction}
\label{sec:intro}
% \CF{one paragraph mentioning importance of energy forecasting and solar forecasting. let Latimar colleagues help, maybe don't mention renewable increase, etc.}

Intra-hour power forecasts are a critical input for reliable and cost-effective grid operation. They enable better dispatch and reserve procurement decisions, support higher-value market participation by asset owners, and reduce the system-wide cost of managing short-term variability through ancillary services, such as regulation reserves. Photovoltaic (PV) generation is especially difficult to forecast at these horizons because cloud motion and localized atmospheric dynamics can sharply shift the plant output in minutes~\cite{barbieri2017}. Purely data-driven models often lack physical grounding for these dynamics, while physics-based models struggle to track fast-moving transient cloud conditions.

% \CF{one paragraph mentioning importance of probabilistic forecasting vs. deterministic forecasting}
% PJM down regulation reserve could go here.
Probabilistic solar forecasts are increasingly recognized as necessary for the market and for grid operation decisions that a deterministic forecast alone cannot support~\cite{li2020review}. Probabilistic forecasts quantify a range of likely outcomes, allowing reserve, bidding, and real-time operating decisions to be sized according to confidence rather than expected output alone by using deterministic forecasting models. This distinction is especially important for services such as PJM's down-regulation reserve, where a deterministic forecast only indicates the expected average flexibility over the interval, while a probabilistic forecast reveals whether that flexibility is likely to be sustained across the full commitment period.

% \CF{one paragraph mentioning current gaps in probabilistic forecasting using ML SORA}
%Recent machine learning approaches extend this probabilistic framework with a range of architectures, from quantile-regression convolutional-recurrent networks to attention-based models combining multi-scale temporal and meteorological context~\cite{wang2026}, and many already integrate site-level measurements, clear-sky modeling, and external irradiance forecasts~\cite{schubnel2025}. The gap lies in exploiting real-time, on-site measurements, particularly inverter-level data, to capture near-term changes in cloud cover and plant state that these models hold static, for lack of access to this kind of telemetry. This is the gap this work addresses.%

Recent machine learning approaches have advanced probabilistic solar forecasting along several fronts, but they typically fall short in three specific ways that this work addresses. First, many remain limited in modality: Attention-based models have combined multiscale temporal history with meteorological context~\cite{wang2026} but without a spatial, inverter-level view of cloud dynamics across the plant. Second, when multiple data sources are available, they are often combined through a single fusion step rather than staged according to each modality's structure; one recent approach fuses sky images and time-series data via a single gated-attention module~\cite{jiang2026}, without staging fusion across more than two modalities or input types. Third, inverter-level telemetry has been used to infer subplant shading and cloud motion but only at second-level granularity using a simple feed-forward network~\cite{meng2021}, without integration into a broader multimodal, intra-hour probabilistic framework. 

% \CF{thinking the innovations of the algorithm: 1. multi-modal learning, 2. inverter-informed (but need to make sure it has positive contribution), 3. deterministic vs. probabilistic (flexible on loss function), whatelse?}
%In this paper, a hierarchical, multi-modal deep learning framework is developed for probabilistic intra-hour solar forecasting. The frameworks takes advantage of existing solar plant and weather data sources typically used in solar resource forecasting today. A deterministic irradiance-to-power baseline is first constructed using a pvlib Ineichen clear-sky model, PySAM DC/AC conversion, and a machine-learning bias correction. Site-level forecasts are then produced by a multi-branch network that separately encodes plant-level temporal history, inverter-level spatial cloud dynamics through a graph-attention encoder, and known-future solar geometry, fused via gated cross-attention and decoded through a smart-persistence-anchored head that supports both deterministic and probabilistic outputs under a flexible loss formulation. Knowledge of inverter topology delivers a consistent improvement over the no-inverter baseline across all sky regimes, including overcast conditions, where spatial information has historically been unreliable. Multi-source fusion with external vendor forecasts, together with sequential online conformal calibration, further refines the probabilistic predictions.\cite{mol2025}%

This work directly addresses and bridges these three gaps by introducing a novel inverter-informed multimodal transformer model for PV generation forecasting, named IMPACT, which fuses a plant's inverter-level telemetry, temporal history, solar geometry, and meteorological snapshots through gated cross-attention, anchoring its predictions to a smart persistence baseline via horizon-conditioned corrections.

% In this paper, we bridge the above gaps by developing a hierarchical multi-modal transformer model for probabilistic intra-hour solar power forecasting. Four branches separately encode the plant's recent temporal history using an attention-based sequence encoder, inverter-level spatial cloud dynamics, known-future solar geometry at each forecast horizon, and the current meteorological snapshot, injected as a shared bias across all forecast horizons. These four representations are fused through gated cross-attention, and a decoder anchors its prediction to a smart-persistence baseline, learning a horizon-conditioned correction. Inverter-level telemetry is shown, through branch-composition ablation, to deliver a positive, non-redundant improvement over a no-inverter baseline across all sky regimes, including overcast conditions, where such signals have historically added little value~\cite{mol2025}. The resulting probabilistic forecasts are refined through sequential online conformal calibration~\cite{xu2023}.

\begin{figure*}[t]
\centering
\includegraphics[width=0.9\textwidth]{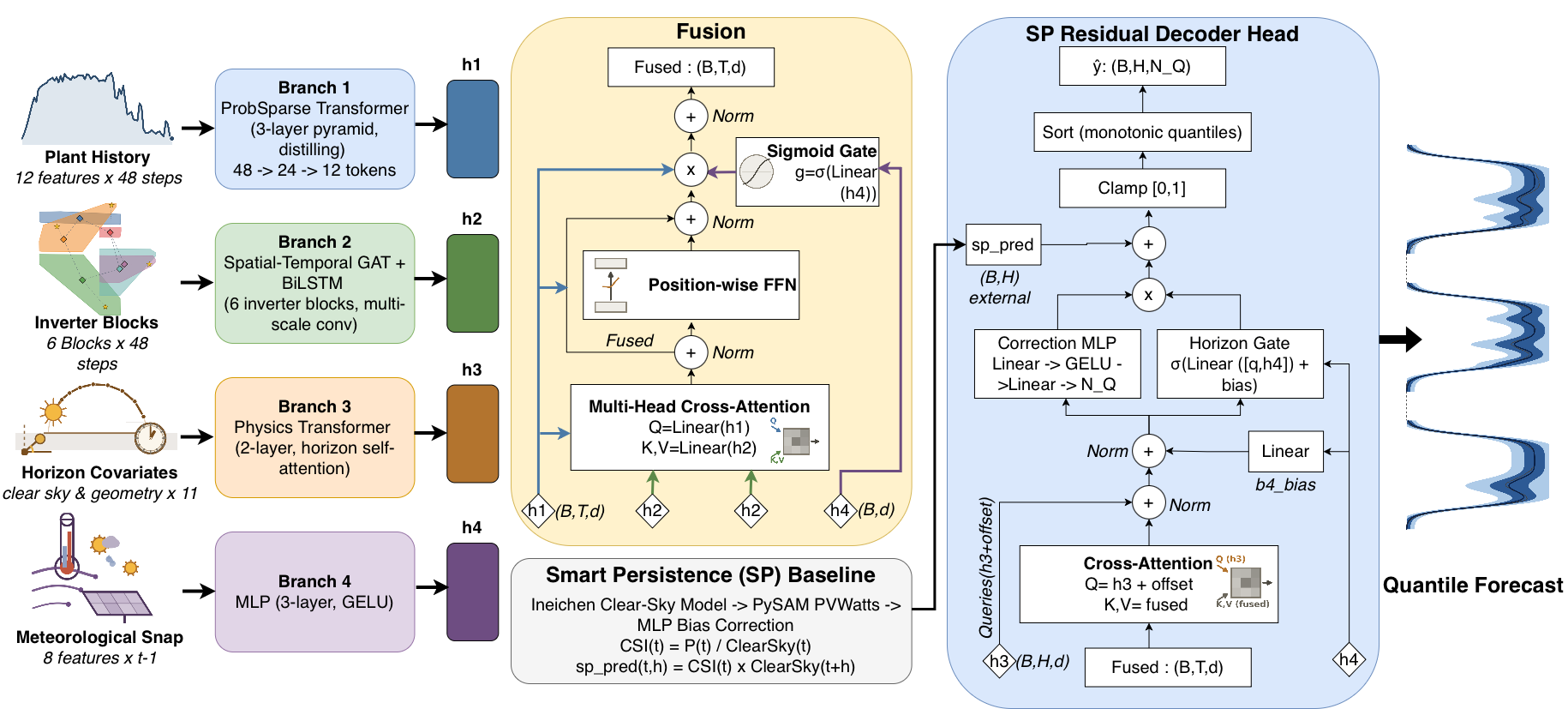}
\caption{Overview of IMPACT framework. Four branch encoders are fused via gated cross-attention and decoded into a per-horizon quantile forecast through a residual head.}
\label{fig:arch}
\end{figure*}

\section{Methodology} \label{methodology}

The IMPACT model predicts a multi-quantile representation of site-level AC power at 11 horizons (10--110 minutes, 10-minute resolution) via a horizon- and quantile-conditioned correction on the smart persistence benchmark (Section~\ref{data}), followed by online calibration. As shown in Fig.~\ref{fig:arch}, four branch encoders feed the network, described in the following sections, followed by fusion and decoding. Calibration (Section~\ref{methodology}-E) is applied afterward.

\subsection{Data}\label{data}
Experiments use data from a 205-MW utility-scale solar plant at a 10-minute resolution from January 2020 through October 2023, including plant-level and per-inverter AC power (112 inverters) and four meteorological station readings of plane-of-array irradiance, global horizontal irradiance, direct normal irradiance, temperature, and wind speed. Plant power is normalized against a clear-sky reference to obtain the clear-sky index (CSI), built from a pvlib Ineichen clear-sky model propagated through a PySAM PVWatts® model, with a walk-forward multilayer perceptron (MLP) bias correction. CSI defines the smart persistence prediction interval (SP-PI) benchmark, used in the evaluations. Sky conditions are classified as clear, variable, and overcast regimes using global horizontal irradiance-based clearness, and samples are grouped by season.
Model development uses three expanding-window splits, each adding 1 year of training data before validating on the following year.

\subsection{Branch Encoders}

Four parallel encoders map heterogeneous input modalities to a shared 96-dimensional embedding space.

\noindent\textbf{Branch 1 (temporal history).} An 8-hour look-back (48 steps, 12 plant-level features: CSI, power, irradiance, and spatial/rolling variability metrics) is encoded by a transformer using ProbSparse self-attention~\cite{zhou2021}, which concentrates capacity away from flat nighttime or clear-sky stretches. Convolutional downsampling between the three encoder layers halves the sequence length at each stage, and a depthwise-separable Conv1d sublayer replaces the standard feed-forward block.

\noindent\textbf{Branch 2 (spatial inverter).} The plant's 112 inverters are aggregated into six blocks following the plant's existing inverter layout in the operational data. Let $\mathrm{CSI}_i(t)$ denote inverter $i$'s raw CSI at time $t$. Each inverter's CSI is first corrected by a fixed per-inverter scale factor, $s_i$, estimated from that inverter's own clear-sky-day median:
\begin{equation}
c_i(t) = \mathrm{CSI}_i(t) / s_i.
\label{eq:inv_corr}
\end{equation}
Block $k$'s mean corrected CSI is then $b_k(t)$, the average of $c_i(t)$ over the inverters in that block, and the anomaly fed to the encoder is:
\begin{equation}
a_k(t) = b_k(t) - \frac{1}{6}\sum_{k'=1}^{6} b_{k'}(t).
\label{eq:anomaly}
\end{equation}
These block-level anomaly sequences pass through multiscale Conv1d layers and a bidirectional long short-term memory (LSTM), then a two-layer graph attention network~\cite{velickovic2018} whose adjacency combines an unconstrained, fully learned pairwise bias term with a separate adaptive term derived from learned node embeddings. A learned gate, a sigmoid of an affine function of the recent cross-block CSI standard deviation, scales this branch's contribution.

\noindent\textbf{Branch 3 (horizon covariates).} At each horizon, a deterministic geometry and calendar vector (cosine solar zenith, sine and cosine of hour of day and day of year) and the origin-time CSI are combined with two horizon-targeted lead features. Let $\sigma_{\mathrm{inv}}(t)$ denote the standard deviation of the CSI across all $N$ inverters at time $t$, a measure of how spatially uneven the sky currently looks. This quantity was found to most strongly correlate with plant-level power volatility roughly 20 minutes later, so it is weighted by a Gaussian of width 40 minutes centered at that lead time:
\begin{equation}
w(h;c) = \exp\!\left(-(h-c)^2 / (2\cdot 40^2)\right),
\label{eq:lead_weight}
\end{equation}
giving the horizon-$h$ feature $\mathrm{spatial\_lead}(t,h) = \sigma_{\mathrm{inv}}(t)\cdot w(h;20)$. Separately, let $\mathrm{ramp\_down\_frac}(t)$ denote the fraction of inverters whose CSI dropped by more than 0.1 over the trailing 20 minutes---a spatial count of local ramps already in progress. This signal was found to lead to a broader ramp-down in the entire plant roughly 10 minutes later, independently of $\sigma_{\mathrm{inv}}(t)$, so it receives the same Gaussian weighting peaked closer to the present, $\mathrm{ramp\_count}(t,h) = \mathrm{ramp\_down\_frac}(t)\cdot w(h;10)$.

\begin{table*}[!t]
\centering
\scriptsize 

\setlength{\tabcolsep}{2pt}
\caption{Deterministic, probabilistic, and per-horizon forecast skill comparison, threefold mean, ranked by FSS. Subscript 4 on MACE/PINAW-avg denotes the mean over the four canonical central interval levels (80/90/95/98\%).}
\label{tab:baselines}
\begin{tabular}{l|cccc|cccc|ccccccccccc}
\toprule
& \multicolumn{4}{c|}{Deterministic} & \multicolumn{4}{c|}{Probabilistic} & \multicolumn{11}{c}{FSS by horizon (min)} \\
\cmidrule(lr){2-5} \cmidrule(lr){6-9} \cmidrule(lr){10-20}
Model & nRMSE & nMAE & $R^2$ & $S_f$ & CRPS & FSS & MACE$_4$ & PINAW-avg$_4$ & 10 & 20 & 30 & 40 & 50 & 60 & 70 & 80 & 90 & 100 & 110 \\
\midrule
\textbf{IMPACT} & \textbf{0.1375} & \textbf{0.0669} & \textbf{0.7683} & \textbf{0.1616} & \textbf{0.0360} & \textbf{0.2950} & \textbf{0.0067} & \textbf{0.2519} & \cellcolor{green!17}0.172 & \cellcolor{green!22}0.221 & \cellcolor{green!26}0.259 & \cellcolor{green!29}0.285 & \cellcolor{green!30}0.304 & \cellcolor{green!32}0.322 & \cellcolor{green!33}0.331 & \cellcolor{green!34}0.338 & \cellcolor{green!34}0.341 & \cellcolor{green!34}0.339 & \cellcolor{green!33}0.334 \\
XGBoost-Q & 0.1452 & 0.0738 & 0.7445 & 0.1146 & 0.0398 & 0.1962 & 0.0072 & 0.3114 & \cellcolor{red!12}-0.242 & \cellcolor{green!5}0.050 & \cellcolor{green!16}0.155 & \cellcolor{green!20}0.204 & \cellcolor{green!24}0.238 & \cellcolor{green!27}0.266 & \cellcolor{green!28}0.281 & \cellcolor{green!29}0.291 & \cellcolor{green!30}0.299 & \cellcolor{green!31}0.310 & \cellcolor{green!31}0.308 \\
QRF & 0.1515 & 0.0803 & 0.7212 & 0.0762 & 0.0422 & 0.1405 & 0.0072 & 0.3747 & \cellcolor{red!20}-0.392 & \cellcolor{red!2}-0.038 & \cellcolor{green!8}0.083 & \cellcolor{green!14}0.144 & \cellcolor{green!19}0.188 & \cellcolor{green!23}0.226 & \cellcolor{green!25}0.247 & \cellcolor{green!26}0.259 & \cellcolor{green!27}0.268 & \cellcolor{green!28}0.280 & \cellcolor{green!28}0.281 \\
Informer-Q & 0.1556 & 0.0787 & 0.7073 & 0.0512 & 0.0424 & 0.1366 & 0.0072 & 0.2951 & \cellcolor{red!22}-0.431 & \cellcolor{red!2}-0.032 & \cellcolor{green!11}0.110 & \cellcolor{green!18}0.179 & \cellcolor{green!22}0.215 & \cellcolor{green!24}0.238 & \cellcolor{green!24}0.244 & \cellcolor{green!25}0.250 & \cellcolor{green!25}0.247 & \cellcolor{green!25}0.248 & \cellcolor{green!24}0.235 \\
TFT-Q & 0.1553 & 0.0817 & 0.7081 & 0.0530 & 0.0436 & 0.1077 & 0.0084 & 0.3291 & \cellcolor{red!27}-0.540 & \cellcolor{red!4}-0.083 & \cellcolor{green!8}0.078 & \cellcolor{green!16}0.163 & \cellcolor{green!20}0.200 & \cellcolor{green!23}0.225 & \cellcolor{green!24}0.237 & \cellcolor{green!24}0.236 & \cellcolor{green!23}0.233 & \cellcolor{green!23}0.227 & \cellcolor{green!21}0.209 \\
SP-PI & 0.1640 & 0.0858 & 0.6674 & 0.0000 & 0.0461 & 0.1058 & 0.0195 & 0.3597 & \cellcolor{green!6}0.062 & \cellcolor{green!8}0.079 & \cellcolor{green!9}0.092 & \cellcolor{green!10}0.102 & \cellcolor{green!11}0.110 & \cellcolor{green!11}0.114 & \cellcolor{green!12}0.118 & \cellcolor{green!12}0.121 & \cellcolor{green!12}0.122 & \cellcolor{green!12}0.122 & \cellcolor{green!12}0.121 \\
LSTM-Q & 0.1593 & 0.0840 & 0.6893 & 0.0287 & 0.0442 & 0.0975 & 0.0128 & 0.3126 & \cellcolor{red!26}-0.519 & \cellcolor{red!4}-0.076 & \cellcolor{green!8}0.081 & \cellcolor{green!15}0.152 & \cellcolor{green!19}0.192 & \cellcolor{green!21}0.214 & \cellcolor{green!22}0.220 & \cellcolor{green!22}0.219 & \cellcolor{green!21}0.210 & \cellcolor{green!20}0.199 & \cellcolor{green!18}0.180 \\
DLinear-Q & 0.1558 & 0.0847 & 0.7060 & 0.0500 & 0.0453 & 0.0904 & 0.0196 & 0.4031 & \cellcolor{red!18}-0.369 & \cellcolor{red!3}-0.062 & \cellcolor{green!5}0.048 & \cellcolor{green!10}0.098 & \cellcolor{green!14}0.135 & \cellcolor{green!16}0.164 & \cellcolor{green!18}0.181 & \cellcolor{green!19}0.191 & \cellcolor{green!20}0.197 & \cellcolor{green!20}0.204 & \cellcolor{green!21}0.205 \\
TCN-Q & 0.1566 & 0.0827 & 0.7043 & 0.0451 & 0.0445 & 0.0819 & 0.0080 & 0.3111 & \cellcolor{red!31}-0.621 & \cellcolor{red!7}-0.138 & \cellcolor{green!4}0.037 & \cellcolor{green!13}0.127 & \cellcolor{green!18}0.178 & \cellcolor{green!21}0.207 & \cellcolor{green!22}0.224 & \cellcolor{green!23}0.229 & \cellcolor{green!23}0.228 & \cellcolor{green!22}0.222 & \cellcolor{green!21}0.207 \\
NLinear-Q & 0.1584 & 0.0855 & 0.6962 & 0.0341 & 0.0462 & 0.0741 & 0.0185 & 0.4056 & \cellcolor{red!19}-0.378 & \cellcolor{red!4}-0.074 & \cellcolor{green!3}0.030 & \cellcolor{green!8}0.082 & \cellcolor{green!12}0.115 & \cellcolor{green!15}0.145 & \cellcolor{green!16}0.162 & \cellcolor{green!17}0.174 & \cellcolor{green!18}0.180 & \cellcolor{green!19}0.188 & \cellcolor{green!19}0.191 \\
GRU-Q & 0.1570 & 0.0857 & 0.6998 & 0.0427 & 0.0453 & 0.0695 & 0.0092 & 0.3095 & \cellcolor{red!30}-0.585 & \cellcolor{red!6}-0.118 & \cellcolor{green!5}0.048 & \cellcolor{green!13}0.129 & \cellcolor{green!17}0.170 & \cellcolor{green!20}0.198 & \cellcolor{green!21}0.205 & \cellcolor{green!20}0.200 & \cellcolor{green!19}0.191 & \cellcolor{green!17}0.174 & \cellcolor{green!15}0.153 \\
SVR-PI & 0.1702 & 0.0931 & 0.6481 & -0.0378 & 0.0535 & -0.0941 & 0.0116 & 0.4160 & \cellcolor{red!40}-0.794 & \cellcolor{red!18}-0.354 & \cellcolor{red!10}-0.202 & \cellcolor{red!4}-0.088 & \cellcolor{red!1}-0.020 & \cellcolor{red!0}-0.001 & \cellcolor{green!4}0.040 & \cellcolor{green!5}0.051 & \cellcolor{green!10}0.103 & \cellcolor{green!12}0.116 & \cellcolor{green!11}0.114 \\
N-BEATS-Q & 0.1726 & 0.1017 & 0.6377 & -0.0524 & 0.0542 & -0.1162 & 0.0093 & 0.3511 & \cellcolor{red!48}-0.952 & \cellcolor{red!17}-0.343 & \cellcolor{red!6}-0.132 & \cellcolor{red!2}-0.030 & \cellcolor{green!3}0.027 & \cellcolor{green!6}0.056 & \cellcolor{green!6}0.064 & \cellcolor{green!5}0.051 & \cellcolor{green!3}0.026 & \cellcolor{red!0}-0.003 & \cellcolor{red!2}-0.043 \\
ARIMA-PI & 0.1739 & 0.1046 & 0.6306 & -0.0604 & 0.0556 & -0.1205 & 0.0085 & 0.4679 & \cellcolor{red!31}-0.620 & \cellcolor{red!16}-0.321 & \cellcolor{red!10}-0.213 & \cellcolor{red!8}-0.149 & \cellcolor{red!5}-0.098 & \cellcolor{red!2}-0.049 & \cellcolor{red!0}-0.014 & \cellcolor{green!1}0.011 & \cellcolor{green!3}0.028 & \cellcolor{green!5}0.046 & \cellcolor{green!6}0.055 \\
PatchTST-Q & 0.1758 & 0.1055 & 0.6278 & -0.0720 & 0.0551 & -0.1397 & 0.0086 & 0.3453 & \cellcolor{red!50}-1.049 & \cellcolor{red!20}-0.409 & \cellcolor{red!9}-0.177 & \cellcolor{red!3}-0.064 & \cellcolor{red!0}-0.004 & \cellcolor{green!3}0.034 & \cellcolor{green!5}0.045 & \cellcolor{green!4}0.043 & \cellcolor{green!3}0.030 & \cellcolor{green!2}0.016 & \cellcolor{red!0}-0.002 \\
\bottomrule
\end{tabular}
\end{table*}

\noindent\textbf{Branch 4 (meteorological snapshot).} A three-layer MLP encodes the most recently measured state: module and ambient temperature, wind speed, instantaneous plane-of-array spatial variability, and beam fraction. Three additional descriptors computed from the full inverter matrix are appended: $\sigma_{\mathrm{inv}}(t)$ itself and the 20- and 60-minute changes in block-to-block CSI spread, which track cloud edges entering or leaving the plant. The two spread-change terms are scaled by $\sigma_{\mathrm{inv}}(t)$, which drives them toward zero during spatially uniform conditions. 

\subsection{Fusion}

Branches 1, 2, and 4 are combined through a staged fusion; Branch 3 bypasses fusion. This staged design addresses the single-gate limitation of prior multimodal PV fusion noted in Section~\ref{sec:intro}~\cite{jiang2026} by combining branches according to each one's input structure rather than through a single fusion step. Branches 1 and 2 are both sequences---over time and over spatial blocks, respectively---so they naturally combine through cross-attention. Branch 4 is a single instantaneous snapshot with no sequence dimension of its own, so it is better suited to modulating the combined signal through a gate. Branch 3 is indexed by forecast horizon and is entirely deferred to the decoder, where it directly initializes the per-horizon query (Section~\ref{decoding}). Let $h_1$ denote Branch 1's temporal sequence and $h_2$ Branch 2's block embeddings. Branch 1 first cross-attends to Branch 2:
\begin{equation}
h_1' = \mathrm{LayerNorm}\big(h_1 + \mathrm{Attn}(Q{=}h_1, K{=}h_2, V{=}h_2)\big),
\label{eq:fusion_attn}
\end{equation}
followed by a position-wise feed-forward layer. Branch 4's embedding, $h_4$, then produces a gate:
\begin{equation}
g = \sigma(W_g h_4 + b_g), \quad g \in (0,1)^d,
\label{eq:fusion_gate}
\end{equation}
and the fused representation blends the spatially enriched sequence with the original temporal one:
\begin{equation}
h_{\mathrm{fused}} = \mathrm{LayerNorm}\big(g \odot h_1' + (1-g) \odot h_1\big).
\label{eq:fusion_blend}
\end{equation}

\subsection{Decoding and Probabilistic Output}\label{decoding}

The decoder predicts a correction on top of the smart persistence benchmark for 99 quantile levels spanning from 0.01 to 0.99. The Branch 3 horizon queries, offset by a learned per-horizon embedding, cross-attend to the fused sequence, receive an additive Branch 4 bias, and are mapped to 99 corrections by a small MLP. A horizon-conditioned gate, shared across quantiles and bounded in $(0,1)$, scales the correction before being added to the persistence anchor and clipped to the valid power range, after which the 99 outputs are sorted along the quantile axis.

\subsection{Loss and Online Calibration}
The network is trained with an equal blend of pinball loss and a continuous ranked probability score (CRPS) score~\cite{feng2025deep}. Because raw network outputs under-cover relative to their nominal quantile levels, a post hoc calibration step corrects them online, separately for each horizon and quantile. Past residuals are stored in a rolling buffer together with a feature vector describing the current conditions (time of day, solar zenith, recent CSI variability). Each buffered residual is then weighted by its similarity to the current feature vector, $x_t$:
\begin{equation}
w_i = \exp\!\left(-\lVert x_t - x_i \rVert^2 / (2\sigma^2)\right),
\label{eq:kernel_weight}
\end{equation}
with weights normalized to sum to 1, and the correction is the weighted quantile of the buffered residuals at the nominal quantile level. This kernel-weighted residual retrieval builds on sequential predictive conformal inference~\cite{xu2023}.

\section{Experimental Setup} \label{expsetup}

\subsection{Evaluation Metrics}

IMPACT's deterministic accuracy against SP-PI uses normalized root mean square error (nRMSE), normalized mean absolute error (nMAE), and $R^2$, computed from each model's median ($q=0.50$) quantile prediction. Forecast skill relative to SP-PI is $S_f = 1 - \mathrm{RMSE}_{\mathrm{model}} / \mathrm{RMSE}_{\mathrm{SP\text{-}PI}}$.
All other comparisons use probabilistic metrics computed post-calibration (Section~\ref{methodology}-E): CRPS and an analogous forecast skill score, $\mathrm{FSS} = 1 - \mathrm{CRPS}_{\mathrm{model}} / \mathrm{CRPS}_{\mathrm{SP\text{-}PI}}$, both integrated over the full 99-point predicted quantile grid and therefore already reflecting calibration and sharpness across the entire predicted distribution. Prediction interval coverage probability (PICP) and prediction interval normalized average width (PINAW)~\cite{paletta2023advances}, by contrast, are each defined at a single nominal level; rather than averaging across all 49 central intervals the 99-point grid admits, we evaluate them at four representative levels (80/90/95/98\%), which are then aggregated as the mean absolute coverage error (MACE, mean $|\mathrm{PICP}_\ell-\ell|$) and PINAW-avg (mean PINAW), both across those levels.
$\mathrm{CRPS}_{\mathrm{SP\text{-}PI}}$ is the uncalibrated benchmark of Section~\ref{data}, so SP-PI's calibrated row reflects the skill gained from the calibration alone, not zero. $S_f$ needs no such caveat because it is computed before the calibration.

\subsection{Baselines}

Fourteen baselines are evaluated on identical data splits, preprocessing, and loss functions. The baseline suite is chosen to span the major forecasting paradigms. Smart persistence is the domain-standard floor, statistical and classical methods represent feature-engineered approaches, linear models test whether the attention's added complexity is required, and deep learning and transformer baselines represent current sequence-forecasting approaches.
\begin{itemize}
    \item Smart persistence: SP-PI, the primary benchmark for all skill scores
    \item Statistical: autoregressive integrated moving average (ARIMA) with automatic order selection
    \item Classical machine learning: support vector regression (SVR), quantile random forest (QRF), XGBoost
    \item Linear: DLinear, NLinear
    \item Deep learning: LSTM, GRU, TCN~\cite{bai2018}, N-BEATS%~\cite{oreshkin2020}
    \item Transformers: Informer~\cite{zhou2021}, TFT~\cite{lim2021}, PatchTST~\cite{nie2023}.
\end{itemize}
All deep learning, linear, and transformer baselines use AdamW with standard regularization and early stopping.

\subsection{Ablation Study}

The contribution of the spatial-inverter and horizon-covariate branches is isolated through a cumulative branch-composition ablation, adding one branch at a time (B1, B1+B2, B1+B2+B3, full model) across all cross-validation folds, with each added branch retaining its full mechanism, as described in Section~\ref{methodology}.

\section{Results} \label{results}

Results are evaluated across multiple angles: skill relative to SP-PI and a broad baseline suite; calibration and sharpness across the full predicted distribution; and the contribution of each architectural component, each further broken down by sky condition, season, and time of day.

\subsection{Deterministic Forecasting Performance}

Table~\ref{tab:baselines} compares IMPACT's median forecast against SP-PI and all 13 baselines averaged across all 11 horizons and 3 cross-validation folds. The model reduces nRMSE from SP-PI's 0.1640 to 0.1375 and nMAE from 0.0858 to 0.0669, increasing $R^2$ from 0.6674 to 0.7683, for an overall forecast skill of $S_f = 0.1616$. It achieves the lowest nRMSE and nMAE and the highest $R^2$ of any model evaluated, leading the strongest deterministic baseline, XGBoost-Q, by 0.008 nRMSE and 0.024 $R^2$.

\subsection{Probabilistic Forecasting Performance}

\begin{figure}[t]
  \centering
  \includegraphics[width=0.8\columnwidth]{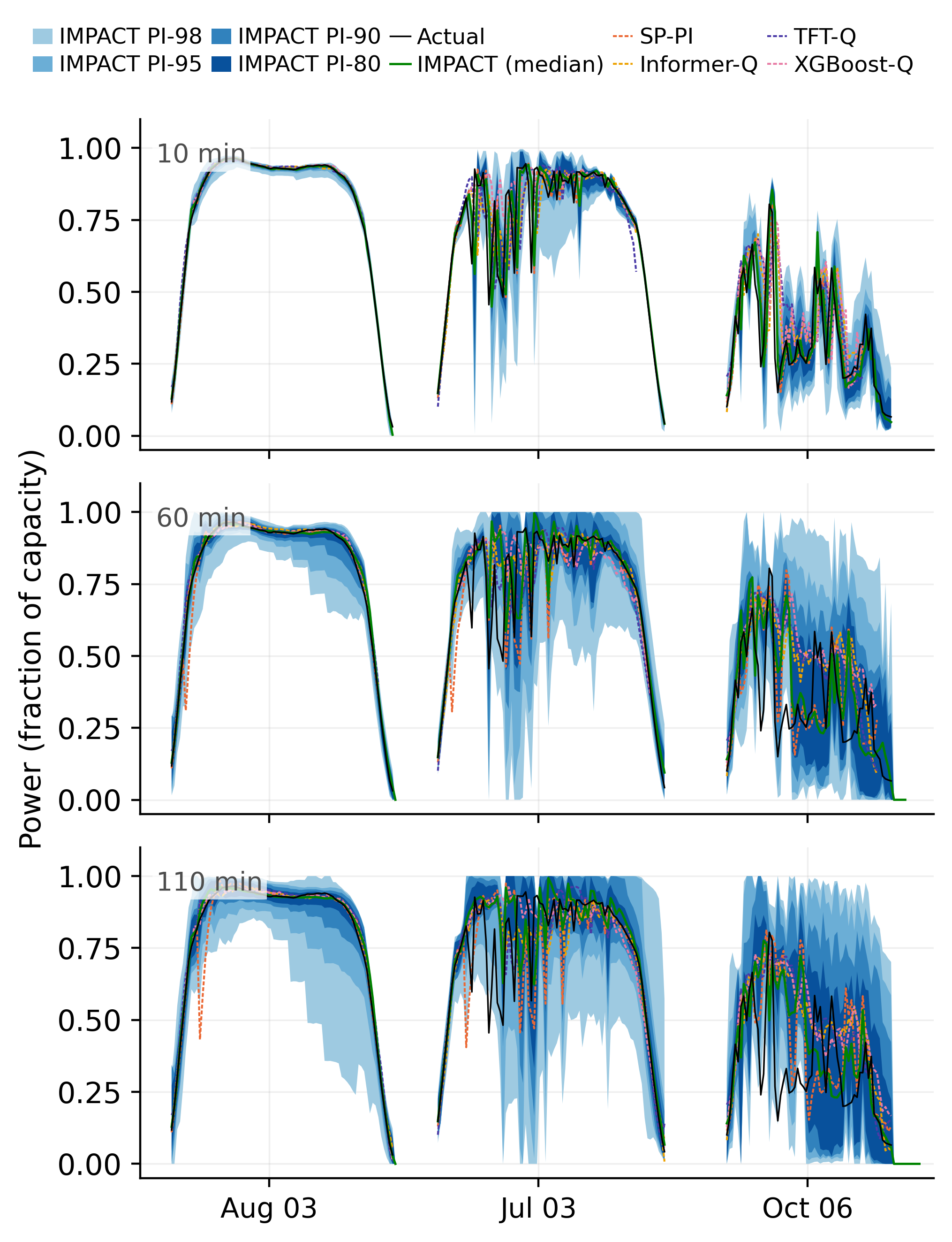}
  \caption{Forecast comparison at three horizons. Each row shows the actual power
  (black) against IMPACT (green, solid) with four nested prediction
  intervals in one blue family (80/90/95/98\%, darkest to lightest) and four
  baselines (dotted) at 10, 60, and 110 minutes across three example days
  spanning clear, partly cloudy, and turbulent conditions.}
  \label{fig:baseline_horizons}
\end{figure}

\begin{figure}[t]
  \centering
  \includegraphics[width=1\columnwidth]{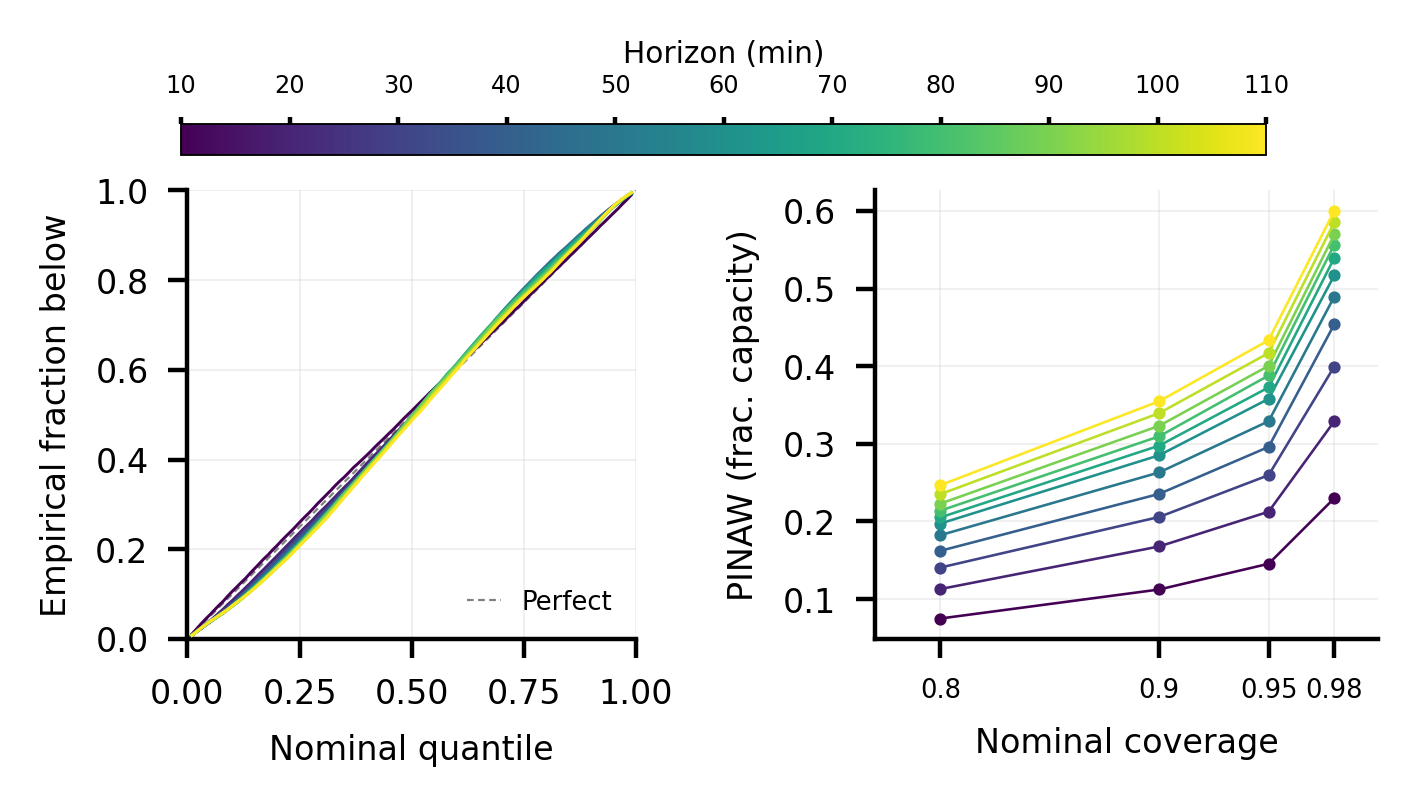}
  \caption{Per-quantile reliability and sharpness of IMPACT's forecasts across all horizons. Left: quantile coverage, with the diagonal denoting perfect calibration. Right: mean PINAW at the four nominal levels (80/90/95/98\%).}
  \label{fig:reliability_sharpness}
\end{figure}

Table~\ref{tab:baselines} ranks IMPACT against all 14 baselines by FSS, leading the strongest (XGBoost-Q) by 0.099 FSS. QRF and Informer-Q form a clear second tier (FSS 0.141 and 0.137), while most of the remaining deep learning and linear baselines cluster more loosely between 0.07 and 0.11 FSS. On MACE, IMPACT (0.0067) is the best-calibrated of all 15 compared models, narrowly ahead of XGBoost-Q and Informer-Q (0.0072 each), and it remains the sharpest on PINAW-avg (0.2519) as well. 
The classical statistical baselines (ARIMA-PI, SVR-PI) fall below SP-PI, indicating that a poorly calibrated model can be actively worse than doing nothing. At 10 minutes, every learned baseline scores below zero, meaning SP-PI alone beats every learned model at the shortest lead time; most baselines do not turn positive until 30 minutes. IMPACT is the only one with positive skill at every horizon, and it remains at the top of the ranking throughout the full 10- to 110-minute range.
Fig.~\ref{fig:baseline_horizons} confirms this pattern visually: At 10 minutes, the baselines' forecasts track actual power far less reliably than IMPACT's, consistent with their negative FSS at that horizon, while by 60 and 110 minutes, the gap narrows as the baselines catch up. 
Fig.~\ref{fig:reliability_sharpness} shows IMPACT's per-quantile reliability and sharpness remain strong across all horizons.

%\subsection{Branch-Composition Ablation}

\subsection{Sensitivity Analysis}

A branch-composition ablation (Table~\ref{tab:ablation}) isolates the contribution of each branch, adding one at a time with its full mechanism. Branch variants were trained and evaluated at the 9-quantile configuration [0.01, 0.025, 0.05, 0.10, 0.50, 0.90, 0.95, 0.975, 0.99], which contains exactly the levels needed to construct the four operationally relevant prediction intervals (80/90/95/98\%) reported throughout this evaluation. Every added branch improves the forecast skill, and the improvement is not evenly distributed: Branch 3's horizon-targeted spatial feature contributes the largest single gain (+0.038 FSS), with others contributing smaller, still positive increments.

\begin{table}[!t]
\centering
\caption{Branch-composition ablation, threefold mean, evaluated at the 9-quantile configuration.}
\label{tab:ablation}
\begin{tabular}{lc}
\toprule
Configuration & FSS \\
\midrule
B1 & 0.2735 \\
B1 + B2 & 0.2920 \\
B1 + B2 + B3 & 0.3304 \\
B1 + B2 + B3 + B4 (full) & 0.3485 \\
\bottomrule
\end{tabular}
\end{table}

A stratified analysis (Table~\ref{tab:stratified}) further shows that the performance systematically varies by sky condition, season, and time of day. By sky regime, skill is highest in clear conditions and lowest in overcast sky, with variable between the two; overcast and variable sky show comparably weak calibration (MACE 0.023 and 0.022) and comparably wide intervals. By season, summer leads clearly, while others are closely grouped behind it. By time of day, both transition windows outperform midday: morning has the widest intervals of the three, while evening is the sharpest; midday is nonetheless the best calibrated (MACE 0.007), narrowly ahead of evening (0.008) and clearly ahead of morning (0.012).

\begin{table}[!t]
\centering
\caption{Stratified analysis, threefold mean.}
\label{tab:stratified}
\begin{tabular}{llcccc}
\toprule
Type & Stratum & CRPS & FSS & MACE & PINAW-avg \\
\midrule
Regime & Clear     & 0.0268 & 0.337 & 0.006 & 0.202 \\
Regime & Overcast  & 0.1103 & 0.198 & 0.023 & 0.646 \\
Regime & Variable  & 0.1094 & 0.255 & 0.022 & 0.654 \\
\addlinespace
Season & Summer & 0.0222 & 0.366 & 0.009 & 0.178 \\
Season & Fall   & 0.0287 & 0.291 & 0.032 & 0.187 \\
Season & Winter & 0.0607 & 0.280 & 0.008 & 0.397 \\
Season & Spring & 0.0615 & 0.280 & 0.021 & 0.403 \\
\addlinespace
Time of day & Morning & 0.0466 & 0.473 & 0.012 & 0.342 \\
Time of day & Midday  & 0.0357 & 0.238 & 0.007 & 0.239 \\
Time of day & Evening & 0.0277 & 0.274 & 0.008 & 0.216 \\
\bottomrule
\end{tabular}
\end{table}

\section{Conclusion and Future Work}\label{sec-conclusion}

This paper presented IMPACT, a multimodal deep learning framework for probabilistic intra-hour PV power forecasting, anchored to smart persistence and refined through online conformal calibration. IMPACT outperformed all 14 baselines, leading the strongest one by 0.099 FSS. Branch-composition ablation showed every branch contributing positive, nonredundant skill, with the spatial and horizon-covariate branches contributing the most. Performance varied by sky regime, season, and time of day, with overcast conditions the hardest and variable close behind, and unexplained asymmetry in calibration behavior between morning and evening transitions.

Future work includes combining IMPACT's forecasts with external vendor irradiance products to extend accuracy across the full horizon range. Access to true geographic coordinates for each inverter could introduce a physically grounded adjacency prior, which the current fully data-driven adjacency lacks. The performance gaps and transition window asymmetry noted above also warrant further investigation.

%References
\bibliographystyle{IEEEtran}
\bibliography{MISO_LF}

\end{document}